\documentclass[conference]{IEEEtran}
\IEEEoverridecommandlockouts

\usepackage{cite}
\usepackage{amsmath,amssymb,amsfonts}
\usepackage{algorithmic}
\usepackage{graphicx}
\usepackage{textcomp}
\usepackage{multirow}
\usepackage{bbm}
\usepackage{mathrsfs}
\usepackage{xcolor}
\usepackage{booktabs}
\usepackage{amsmath}
\usepackage{amssymb}
\usepackage{mathtools}
\usepackage{amsthm}
\usepackage{xspace}
\usepackage{bm}  
\usepackage[utf8]{inputenc}
\usepackage{algorithm,multirow,hyperref}
\usepackage{enumitem}

\newcommand{\fboost}{\textsc{Fair MP-Boost}\xspace}
\newcommand{\fis}{\emph{TreeFIS}\xspace}
\newcommand{\fairfis}{\emph{FairTreeFIS}\xspace}

\newcommand{\by}{\bm{y}}
\newcommand{\bz}{\bm{z}}

\newcommand{\bX}{\bm{X}}

\newcommand{\bp}{\bm{p}}
\newcommand{\bq}{\bm{q}}

\newcommand{\Hcal}{\mathcal{H}}

\newcommand{\ones}{{{\mathbbm{1}}}}
\newcommand{\ind}[1]{\ones_{\{#1\}}}
\newtheorem{definition}{Definition}

\def\BibTeX{{\rm B\kern-.05em{\sc i\kern-.025em b}\kern-.08em
    T\kern-.1667em\lower.7ex\hbox{E}\kern-.125emX}}
\begin{document}

\title{Fair MP-BOOST: Fair and Interpretable Minipatch Boosting\\

\thanks{COL acknowledges support from the NSF Graduate Research Fellowship Program under grant number 1842494.  GIA and COL acknowledge support from the JP Morgan Faculty Research Awards and NSF
DMS-2210837.}
}

\author{\IEEEauthorblockN{Camille Olivia Little}
\IEEEauthorblockA{\textit{Department of Electrical and Computer Engineering} \\
\textit{Rice University}\\
Houston, Texas \\
col1@rice.edu}
\and
\IEEEauthorblockN{Genevera I. Allen}
\IEEEauthorblockA{\textit{Department of Statistics} \\
\textit{Columbia University}\\
New York, NY \\
genevera.allen@columbia.edu}
}

\maketitle

\begin{abstract}
Ensemble methods, particularly boosting, have established themselves as highly effective and widely embraced machine learning techniques for tabular data. In this paper, we aim to leverage the robust predictive power of traditional boosting methods while enhancing fairness and interpretability. To achieve this, we develop \fboost, a stochastic boosting scheme that balances fairness and accuracy by adaptively learning features and observations during training.  Specifically, \fboost sequentially samples small subsets of observations and features, termed minipatches (MP), according to adaptively learned feature and observation sampling probabilities.  We devise these probabilities by combining loss functions, or by combining feature importance scores to address accuracy and fairness simultaneously.  Hence, \fboost prioritizes important and fair features along with challenging instances, to select the most relevant minipatches for learning. The learned probability distributions also yield intrinsic interpretations of feature importance and important observations in \fboost. Through empirical evaluation of simulated and benchmark datasets, we showcase the interpretability, accuracy, and fairness of \fboost.
\end{abstract}



\section{Introduction}
Machine learning (ML) systems have infiltrated society in numerous industries including healthcare, finance, and education \cite{Pessach:2022}. These models undoubtedly add value in decision-making as they can leverage large amounts of information and help scale processes. Ensemble and deep learning systems are amongst the most popular predictive ML methods due to their strong predictive accuracy \cite{das:2017}. While these methods optimize for accuracy, most lack interpretability, or the ability for a human to understand how the model produced its results. Furthermore, with the growing use of ML in high-stakes applications, there has been concern and evidence of bias within existing systems \cite{kleinberg:2018}. The combination of many popular models lacking both interpretability and fairness leads to mistrust amongst users. 

Several methods have addressed bias in ML systems \cite{Pessach:2022}. Specifically, some methods attempt to improve fairness by rebalancing the dataset concerning underrepresented groups, removing features that may be correlated with the protected attribute, or adding synthetic data to increase representation \cite{navarro:2023}. Others add a fairness-aware penalty or constraint to mitigate bias \cite{cruz:2022}. While these methods effectively enhance fairness, they often neglect the crucial intersection of fairness and interpretability. 

While interpretability in machine learning, particularly in discerning important features, has been extensively explored \cite{allen:2023}, its application to fairness or boosting methods remains relatively underexplored. Traditional boosting techniques like AdaBoost and gradient boosting excel in predictive accuracy but typically overlook bias mitigation. Although fair boosting methods exist, they often lack interpretability regarding fairness \cite{grari:2019, cruz:2022}. Our objective is to devise an interpretable approach that achieves high performance in both accuracy and fairness.

Inspired by both MP-Boost which offers feature interpretability \cite{Toghani:2021} and fair adversarial boosting \cite{grari:2019} which offers improvements in fairness, we propose a novel algorithm \fboost which is a double stochastic gradient boosting technique based upon minipatches (MPs) \cite{Yao:2020}. Our algorithm adaptively learns MPs which are tiny subsets of features and observations, at each iteration that balance fairness and accuracy for both observations and features. Our learning scheme allows us to leverage traditional feature importance scores for trees like the MDI \cite{breiman:1973} and the recently proposed \fairfis \cite{little:2023}. These learned feature probability distributions allow practitioners to interpret the predictions of \fboost. The adaptive sampling of observations further ensures the most relevant minipatches are being sampled for training. 

\textbf{Contributions.} 
In this work, we develop a fair and interpretable boosting scheme. Our contributions consist of developing a novel algorithm with the major novel components being an adaptive learning scheme for observation and feature probabilities that balance fairness and accuracy. The result of this is an interpretable boosting model that allows us to understand which features and observations are useful for balancing fairness and accuracy. Through empirical analysis, we validate interpretations of our feature subsampling scheme and we demonstrate that \fboost outperforms state-of-the-art bias mitigation algorithms.

\label{sec:background}
\section{Fair MP-Boost}
\label{sec:fboost}
Motivated by MP-Boost \cite{Toghani:2021}, we aim to devise a boosting approach that maintains fairness and interpretability without compromising accuracy. We begin by defining group bias metrics and introducing helpful notation. Let $z_i \in \{0,1\}$ for $i \dots n$ denote an indicator variable for the protected attribute (e.g., gender, race). Group fairness entails uniform predictions across all demographic groups \cite{Beutel:2017}. Specifically, Demographic Parity (DP) assesses whether predictions $\hat{\by}$ are independent of the protected attribute $\bz$. We formalize DP as follows.

\begin{definition}
\textbf{Demographic Parity} A classifier exhibits fairness if the prediction $\mathbf{\hat{y}}$ is independent of the protected attribute $\mathbf{z}$.
\begin{equation}
P(\mathbf{\hat{y}} = 1| \mathbf{z} = 1) = P(\mathbf{\hat{y}} =1 | \mathbf{z} = 0 )
\end{equation}
\end{definition}

While we focus on implementing DP in this study, our approach readily extends to other fairness criteria, such as Equality of Opportunity \cite{Beutel:2017}. We represent our dataset as $(\bX, \by, \bz) \in (\mathbb{R}^{N \times M}, \mathbb{R}^N, \mathbb{R}^N)$. Concentrating on binary classification tasks, each sample $x_i \in \mathbb{R}^M$ corresponds to a label $y_i \in \{-1,+1\}$ and a protected attribute $z_i \in \{0,1\}$. 

\subsection{Algorithm Framework}
Our primary goal is to develop a classifier that is fair, accurate, and interpretable. Our key idea to do this is to leverage a double stochastic gradient boosting technique where we adaptively learn both features and observations as the algorithm proceeds. We propose to do this by utilizing minipatches, where we adaptively sample features and observations that contribute to fairness and accuracy \cite{Yao:2020}. Formally, a minipatch is defined as $(\bX_{\mathcal{R},\mathcal{C}},\by_{\mathcal{R}})$, where $\mathcal{R}$ denotes a subset of examples with size $n$, and $\mathcal{C}$ denotes a subset of features with size $m$. Furthermore, we define $\Hcal$ to be the class of decision trees, where each $h \in \Hcal$ is a function $h: \mathbb{R} \rightarrow \{-1, +1\}$.

The main novel components of our algorithm are the adaptively learned observation and feature sampling schemes. As outlined in Algorithm 1, we first sample an MP based on adaptive observation and feature probability distributions $\bp$ and $\bq$ initially set to be uniform $\left(\mathbf{U}_{[N]}\right)$ and $\left(\mathbf{U}_{[M]}\right)$, where $[N] = 1,2 \dots, N$ and $[M] = 1,2, \dots, M$ respectively. Then, we fit a tree to the minipatch, and lastly, we update the output given the ensemble function $E: \mathbb{R}^M \rightarrow \mathbb{R}$. The inspiration for this algorithm structure is taken from MP-Boost \cite{Toghani:2021}, however, the novel component is how we learn and adaptively update the observation probability $\bp$ and the feature probability $\bq$.

Additionally, and similar to other stochastic boosting regimes that utilize out-of-bag samples for internal validation, we propose to leverage out-of-patch samples for internal validation including early stopping and possibly hyperparameter tuning.  We adopt a scheme similar to MP-Boost \cite{Toghani:2021} to compute the out-of-patch error.  Specifically, we aggregate the output of weak learners that do not include sample $i$ it in their minipatch: $G^{(t)}(\mathbf{x_i}) = G^{(t-1)}(\mathbf{x}_i) + h^{(t)}\left(\mathbf{x}_{i_{\mathcal{C}^{t}}}\right) \forall_i \notin \mathcal{R}^{(t)}$. The out-of-patch update is defined as
\begin{align}
\label{eqn:oop}
\text{oop}^{(t)} = \frac{1}{i \notin \mathcal{R}^{(t)}}\sum_{i=1}^{i \notin \mathcal{R}^{(t)}}\ind{sgn(G^{(t)}(\mathbf{x}_i)) = y_i}.
\end{align}
This out-of-patch output serves as a conservative estimate of test accuracy. Our early stopping rule is straightforward: if the current out-of-patch performance increases beyond a certain margin, the algorithm requires more time to refine; otherwise, it is deemed to generalize well to the test set and learning can cease.

\begin{algorithm}
\label{alg:fair_boost}
\textbf{Input} $\bX, \by, \bz,m,n,\mu, \alpha$\\
$\bp^{(0)} = \left(\mathbf{U}_{[N]}\right) $\\
$\bq^{(0)} = \left(\mathbf{U}_{[M]}\right)$\\
  \textbf{While early stopping not met:}
  \begin{enumerate}[leftmargin=*]
  \item \textbf{Subsample Minipatch}: Sample $n$ observations $\mathcal{R}^{(t)} \subset [N]$ with probability $\bp^{(t)}$ \&  Sample $m$ features $\mathcal{C}^{(t)}\subset [M]$ with probability $\bq^{(t)}$.
  \item \textbf{Fit decision tree $h^{(t)}$to minipatch}: $\left(\bX_{\mathcal{R}^{(t)},C^{(t)}},\by_{\mathcal{R}}^{(t)}\right)$.
 \item \textbf{Update Ensemble}: $E^{(t)}(x_i) = E^{(t-1)}(x_i)+ h^{(t)}\left(x_{i_{\mathcal{C}^{(t)}}}\right)$.
 \item \textbf{Update observation probability distributions}: (Eqn \ref{eqn:obs_probs}).
 \item \textbf{Update feature probability distributions}: (Eqn \ref{eqn:feat_probs}).
  \item \textbf{Compute Out-of-Patch Accuracy} (Eqn:~\ref{eqn:oop}): 
  \end{enumerate}
  
\textbf{Output} $sgn(E^{(T)})$

\caption{Fair Minipatch Ensemble Learning}
\label{alg}
\end{algorithm}

\subsection{Adaptive feature and observation sampling}

The main novel component of our \fboost algorithm is the development of a novel adaptive scheme to learn observation and feature sampling probabilities, $\bp$ and $\bq$ respectively.  The key idea of our proposal is that $\bp$ and $\bq$ should be based on a linear combination of functions that trade off accuracy and fairness.  Specifically, we define a hyperparameter, $\alpha$, that controls this tradeoff.  When $\alpha = 0$, our \fboost algorithm solely focuses on accuracy, which reverts to the MP-Boost algorithm introduced by \cite{Toghani:2021} as a special case. When $\alpha = 1$, \fboost solely focuses on fairness and represents a novel fairness-focused boosting algorithm.  When $\alpha \in (0,1)$, \fboost balances both accuracy and fairness. The same tradeoff parameter $\alpha$ is used in both the feature and observation sampling probabilities.

Let us first dive into the specifics of the observation sampling scheme.  The general idea is to prioritize observations that are more difficult to learn. For accuracy, misclassified samples are more difficult to learn, so we propose to increase their probabilities to be sampled more frequently. Formally, let $\mathcal{L}_A: \mathbb{R} \times \mathbb{R} \rightarrow \mathbb{R}^+$ be a function that measures the similarity between the ensemble outputs and labels. To ensure misclassified observations are sampled more, we assign an accuracy-based probability proportional to $\mathcal{L}_A(y_i, E(x_i))$ to the $i^{th}$ observation.

The fairness-focused observation update is a bit trickier because the fairness of a single observation cannot explicitly be measured using common group fairness measures. Inspired by \emph{Fair Adversarial Gradient Tree Boosting (FAB)}\cite{grari:2019}, at each iteration, we propose to use an adversarial classifier $h^{(t)}_F$ that predicts the protected attribute based on the output of $h^{(t)}$. Using the function $h_F:\mathbb{R}\rightarrow \mathbb{R}^+$, observation probabilities are iteratively updated to upsample those with a higher probability of predicting the protected attribute based on $h^{(t)}_F$ outputs. Leveraging $\mathcal{L}_F: \mathbb{R} \times \mathbb{R} \rightarrow \mathbb{R}^+$, which measures the similarity between the protected attribute and the probability of observing it given the adversarial model's output, we update the fairness-based probabilities to be proportional to $\mathcal{L}_F\left(z_i, h_F^{(t)}\left(h^{(t)}\left((x_i)_{C^{(k)}}\right)\right)\right)$ for observations $i \in \mathcal{R}^{(t)}$. Combining both $\mathcal{L}_A$ and $\mathcal{L}_F$, the observation probability update can be defined as:
\begin{align}
\label{eqn:obs_probs}
\tilde{\bp} = (1-\alpha)\mathcal{L}_A+  (\alpha)\mathcal{L}_F.
\end{align}
The exponential or cross-entropy functions are suitable loss functions for $\mathcal{L}_A$ and $\mathcal{L}_F$. As an example and in this paper's empirical studies, we use the exponential loss for both $\mathcal{L}_A$ and $\mathcal{L}_F$. Lastly, we normalize the probabilities to ensure they sum to one: $\bp^{(t+1)} = \tilde{\bp}/\sum_{i=1}^N\tilde{\bp}_i$. 

Inspired by the MP-Boost algorithm which adaptively learns features, we seek to do so in a manner that balances both accuracy and fairness. To this end, we need a measure for each tree of how features contribute to both fairness and accuracy. For accuracy, we rely on what we call \fis (MDI) \cite{breiman:1973}; for fairness, we rely on the recently proposed \fairfis \cite{little:2023} which uses an MDI-like metric based on group fairness. By leveraging \fis and \fairfis, we dynamically select features that optimize both accuracy and fairness and facilitate a clearer interpretation of our ensemble model's outcomes.

Formally, let $\bq$ be the probability distribution of features. Additionally, let $I_j^{h^{(t)}}$ be accuracy feature importance scores computed using \fis (MDI) and $F_j^{h^{(t)}}$ be the fairness feature importance score computed using the \fairfis for features $j \in C^{(t)}$. Different from \fis, \fairfis can be both positive and negative. Positive values correspond to features that contribute to fairness and negative values indicate features contributing to bias. This means that the \fairfis values are not on the same scale as \fis which are non-negative. To account for this, we propose to use a ReLu function $R(x) = max(0,x)$ that zeros out biased features. Furthermore, to ensure we do not establish feature sampling probabilities too early in learning, we include a slow learning term $\mu \in (0,1)$, which determines the ratio of exploration versus exploitation. The feature probability update can be defined as
\begin{align}
\label{eqn:feat_probs}
\tilde{\bq}_j = (1-\mu)q_j^{(t)} + 
\mu \left(\alpha*R(F_j^{h(t))})+ (1 - \alpha)*I_j^{h(t)}\right), 
\end{align}
for every $j \in C^{(t)}$. Note that $\tilde{\bq}_j = max(\tilde{\bq}_j, \epsilon)$. Following inspiration from multi-armed bandits, this ensures that feature sampling probabilities are never set to exactly zero and hence always open for exploration. Analogous to the observation probabilities, we then normalize the probabilities to ensure they sum to one: $\bq^{(t+1)}_j = \tilde{\bq}_j/\sum_{i=1}^N\tilde{\bq}_j$.

Finally, our novel adaptively learned observation and feature sampling probabilities also yield intrinsic interpretations of leverage points and feature importance scores respectively.  Specifically, we propose to average the observation and feature sampling probabilities after a burn-in phase when the probabilities have stabilized. In this paper, our empirical work focuses on validating our feature sampling probabilities as a measure of intrinsic feature importance.

\subsection{Hyperparameter Tuning}
The minipatch size plays a pivotal role as a hyperparameter in our algorithm. Inspiration from boosting suggests having a series of weak learners. If the minipatches are smaller, they are going to be weaker learners, hence $ n = \sqrt{N}$ and $m = \sqrt{M}$. Another critical hyperparameter in our algorithm is the learning rate \(\mu\), for which we set as a default to \(\mu = 0.2\). Additionally, we expect that a user may specify the level of $\alpha$ they want, but if not this tuned using the Fairness-AUC \cite{little:2022}. It is also worth noting that these hyperparameters can be selected in a data-driven manner using our out-of-patch criterion.

\section{Empirical Studies}
\label{sec:empirical}
\begin{table}[htp]
  \caption{Simulation comparative results for various boosting and bias mitigation methods over 10 train/test splits. Our Fair MP-Boost method achieves the highest accuracy and fairness metrics.}
  \label{tab:sim_results}
  \centering
  \begin{tabular}{ccc}
  \toprule
Method                                      & Accuracy    & Fairness    \\
\toprule
Random Forest                                       & 0.89 (.006) & 0.54 (.006) \\
Gradient Boosting                           & 0.90 (.005) & 0.56 (.008) \\
FAB $(\lambda = 0.2)$        & 0.58 (.023) & 0.56 (.022) \\
FAB $(\lambda = 0.05)$        & 0.62 (.021) & 0.59 (.022) \\
Adversarial                   & 0.54 (.043) & 0.29 (.135) \ \\
Fair GBM &                  0.90 (.004) & 0.57 (.005)\ \\
\midrule
FMP-Boost ($ \alpha=0.9$) & 0.82 (.008) & \textbf{0.64 (.007)} \\
FMP-Boost ($\alpha = 0.1$) & \textbf{0.91 (.014)} & 0.57 (.011) 
\end{tabular}
\end{table}\subsection{Simulation Setup and Results}
We design simulation studies to validate our proposed Fair MP-Boost method in terms of interpretability, accuracy, and fairness. We utilize four feature groups: features within $G_1$ and $G_2$ exhibit correlation with the protected attribute $z$ and thus exhibit bias; features within $G_1$ and $G_3$ are indicative features linked to the outcome $y$; and features within $G_4$ solely contribute noise. We simulate the protected attribute, $z_i$, as $\mathbf{z}\stackrel{i.i.d}{\sim} Bernoulli(\pi)$ and take $\pi = 0.2$.  Then, the data is generated as $\mathbf{x}_{i,j}\stackrel{i.i.d}{\sim}N(\alpha_j*z_i, \boldsymbol{\Sigma})$ with $\alpha_j = 2$ if $j \in G_1$ or $G_2$ and $\alpha_j = 0$ if $j \in G_2$ or $G_4$. Hence, all features in $G_1$ and $G_2$ are strongly associated with $z$ and should be identified as biased. Furthermore, we consider a non-linear additive scenario where $f(x_i) = \beta_0 + \sum_{j=1}^p\beta_jsin(x_{ij})$. We also let $\beta_j = 1$ for $j \in G_1$ or $G_3$ and $\beta_j = 0$ for $j \in G_2$ or $G_4$. We employ a logistic model with  $y_i \stackrel{i.i.d}{\sim} Bernoulli (\sigma( f(x_i))$, where $\sigma$ is the sigmoid function.

\begin{figure}[htbp]
\centerline{\includegraphics[scale=0.65]{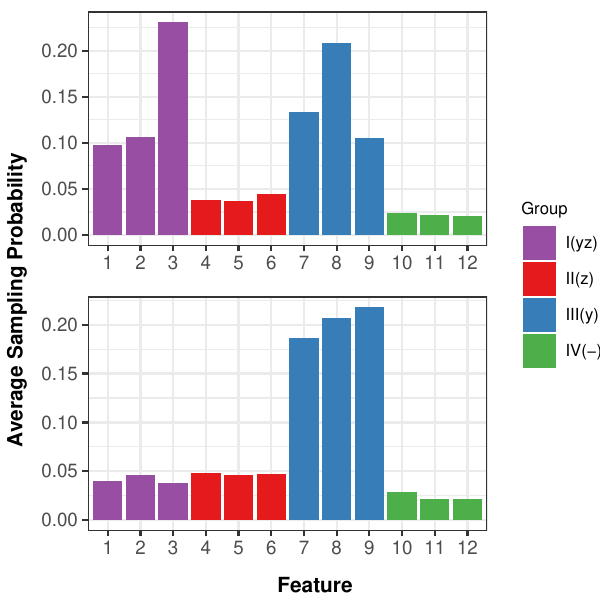}}
\caption{Average feature sampling probabilities after a 20 iteration burn-in phase of Fair MP-Boost on the simulated dataset with $\alpha = 0.1$ (top; accuracy prioritized) and $\alpha = 0.9$ (bottom; fairness prioritized).  Fair MP-Boost correctly learns features associated with the signal (blue and purple) when $\alpha = 0.1$ and features associated with the signal but independent from the protected attribute (blue) when $\alpha = 0.9$; these results validate the feature interpretability of Fair MP-Boost. }
\label{fig:feature_samp_probs}
\end{figure}

We begin by validating the feature sampling probabilities of \fboost. In Figure~\ref{fig:feature_samp_probs}, we present average feature sampling probabilities of \fboost on the simulated dataset with $\alpha = 0.1$ in the top panel and $\alpha = 0.9$ in the bottom panel. In the top panel, when accuracy is prioritized, features in $G_1$ and $G_3$, or the features associated with the response $y$ have higher sampling probabilities than the features correlated with the protected attribute $z$ and the noise features. Note that while features in $G_2$ and $G_4$ have lower sampling probabilities, they are not zeroed out and are still sampled between 3-5\% of the time. Similarly, in the top panel where fairness is prioritized, the features in $G_3$ which are uncorrelated with the protected attribute are sampled the most. Looking at these average feature sampling probabilities allows us to interpret important features and validate our \fboost feature subsampling scheme. 

Additionally, we consider the performance of \fboost to various boosting and bias mitigation strategies. We directly compare our method with two relevant approaches: Fair Adversarial Gradient Tree Boosting (FAB) \cite{grari:2019} and the FairGBM method \cite{cruz:2022}, as well as an Adversarial approach \cite{Zhang:2018}. Furthermore, FAB has a hyperparameter that controls the fairness-accuracy tradeoff, hence we show results for two values. We use default hyperparameter tuning in the available software for all of these approaches. In Table~\ref{tab:sim_results}, we compare Random Forest, Gradient Boosting, FAB, the Adversarial approach, and FairGBM to \fboost and report their accuracy and fairness metrics on the simulated dataset. We observe that while Random Forest, Gradient Boosting, and FairGBM have comparable accuracy, \fboost achieves higher accuracy when accuracy is prioritized ($\alpha = 0.1$). Note also that even when accuracy is prioritized, \fboost achieves equal if not higher fairness than all other methods. When fairness is prioritized ($\alpha = 0.9$), \fboost significantly outperforms all other methods in terms of fairness, while still achieving comparable accuracy in comparison to other methods. While the fairness-accuracy tradeoff is evident in our results, \fboost still outperforms other methods in terms of accuracy given it achieves the highest fairness. These results strongly validate \fboost's performance in terms of accuracy, fairness, and interpretability. 

\subsection{Case Study on Real Data}
\begin{figure*}[htb]
\centerline{\includegraphics[scale=0.55]{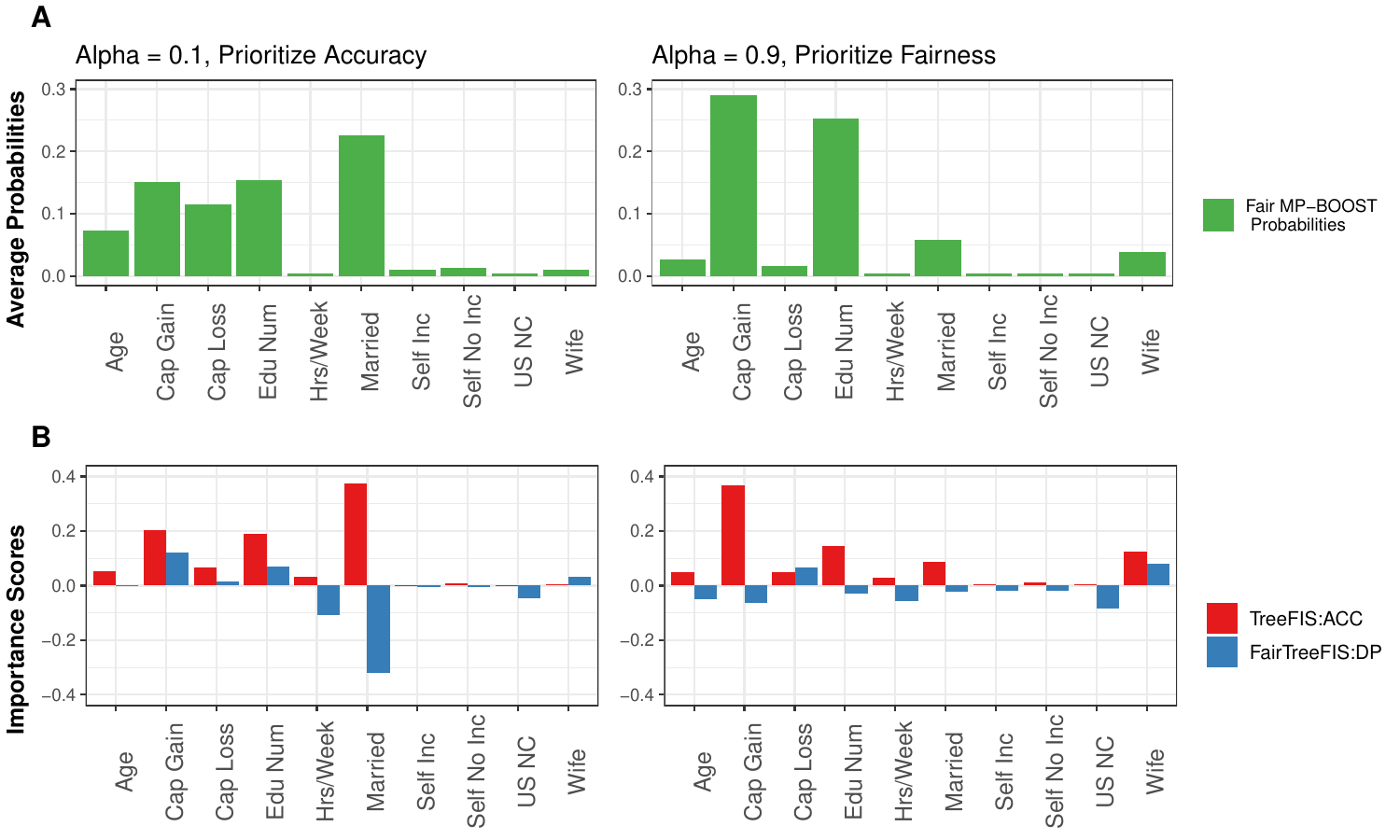}}
\caption{Interpretation of the Adult dataset with Gender as the protected attribute using our \fboost algorithm. Part A shows \fboost average feature sampling probabilities when accuracy is prioritized (left) and when fairness is prioritized (right).  Part B shows feature importance metrics using MDI (TreeFIS) and the fairness-based \fairfis for these same \fboost models.  When accuracy is prioritized, \fboost heavily utilizes the feature \emph{Married}, but this feature also strongly contributes to bias in the model (negative \fairfis score).  When fairness is prioritized, however, \fboost utilizes less of the feature \emph{Married} and more of the features \emph{Capital Gain} and \emph{Edu Num} which leads to improved accuracy and fairness. Overall, the interpretability of our feature sampling probabilities from \fboost aligns with interpretations in Part B, hence validating our results.}
\label{fig:adult_interpretations}
\end{figure*}

\begin{table}[hbt]
\small
  \caption{Real data comparative results for various boosting and bias mitigation methods on the Adult dataset with Gender as the protected attribute and the Law School Admission Bar Passage dataset with Race as the protected attribute over ten train/test splits. Our Fair MP-Boost method achieves the highest accuracy and fairness metrics for both datasets. }
  \label{tab:stackeo}
\centering
\scalebox{0.9}{
  \begin{tabular}{cccccc}
    \toprule
    \multirow{2}{*}{Method} &
      \multicolumn{2}{c}{Adult (Gender)} &
      \multicolumn{2}{c}{Law (Race)} \\
    & Accuracy & Fairness & Accuracy & Fairness \\
    \midrule
    RF & .84(.001) & .82(.001) & \textbf{.89(.002)} & .85(.003)  \ \\
    Gradient Boosting & .85(.030) & .83(.007) & .88(.004) & .86(.052)  \ \\
    FAB (0.2) & .81(.002) & .92(.002) & .85(.003) & .93(.004)  \ \\
    FAB (0.05) & .83(.002) & .86(.002) & .87(.005) & .88(.005) \ \\
    Adversarial & .82(.002) & .91(.017) & \textbf{.89(.001)}     & .84(.019)\ \\
    Fair GBM & \textbf{.86(.001)} & .82(.001) & \textbf{.89(.001)} & .81(.004)\ \\
    \midrule
    FMP-Boost (0.9)& .80(.004) & \textbf{.94(.004)} &  .86(.004) & \textbf{.99(.003)}  \ \\
    FMP-Boost (0.1) & \textbf{.86(.003)} & .87(.003) & \textbf{.89(.002)} &  .92(.004) \ \\
    \bottomrule
  \end{tabular}} \vspace{-0.2in}
\end{table}
To align our work with existing fairness literature, we evaluate our method on two popular benchmark datasets. We examine the Adult Income dataset containing 14 features and approximately 48,000 individuals with class labels stating whether their income is greater than $\$50,000$ and Gender as the protected attribute \cite{du:2019}. We also consider the Law School dataset, which has 8 features and 22,121 law school applicants with the class labels stating whether an individual will pass the Bar exam when finished with law school and Race as the protected attribute \cite{law_data}.

We first interpret our $\fboost$ algorithm on the Adult Dataset with Gender as the protected attribute. Part A of Figure~\ref{fig:adult_interpretations} displays the average feature sampling probabilities for $\fboost$ when prioritizing accuracy (left) and fairness (right). Part B shows feature importance metrics using $\fis$ and $\fairfis$ for fair and unfair models. Notably, when accuracy is prioritized, the feature \emph{Married} is sampled frequently but also exhibits a very negative $\fairfis$ score, indicating its contribution to model bias. Conversely, when prioritizing fairness, \emph{Married} is utilized less. The $\fairfis$ metric (blue bars) facilitates the interpretation of fairness-contributing features, while the red bars show features impacting accuracy. When $\alpha = 0.9$, none of the features significantly contribute to bias in the model. Moreover, \emph{Married} is notably underutilized, resulting in a fairer model. The MDI feature importance (red bars) closely aligns with our sampling probabilities, affirming the validity of using average feature sampling probabilities as an intrinsic metric for feature importance.

Lastly, in Table~\ref{tab:stackeo}, we compare accuracy and fairness results on several boosting and bias mitigation techniques using the Adult Income and Law School datasets. For the Adult dataset, \fboost achieves comparable if not better performance in terms of accuracy in comparison to the other methods. For fairness, \fboost outperforms all other methods. Similarly, for the law school dataset, \fboost achieves comparable accuracy and outperforms other methods in terms of fairness. These results strongly validate our work.

\section{Discussion}
In this work, we introduce an interpretable boosting method aimed at enhancing group fairness while maintaining accuracy. A significant advantage of our approach lies in its interpretability, as it allows direct insight into the features influencing both accuracy and fairness.

\bibliographystyle{IEEEbib}
\bibliography{reference.bib}

\end{document}